\documentclass{article}

\usepackage{arxiv}

\usepackage[utf8]{inputenc} 
\usepackage[T1]{fontenc}    
\usepackage{hyperref}       
\usepackage{url}            
\usepackage{booktabs}       
\usepackage{amsfonts}       
\usepackage{nicefrac}       
\usepackage{microtype}      
\usepackage{graphicx}       
\usepackage{amsmath,amssymb}
\usepackage{adjustbox}
\usepackage{threeparttable}
\usepackage{bm}
\usepackage{float}
\usepackage{xcolor}
\usepackage{color, colortbl}
\usepackage[caption=false,font=footnotesize]{subfig}
\newcolumntype{P}[1]{>{\centering\arraybackslash}p{#1}}
\definecolor{atomictangerine}{rgb}{1.0, 0.7, 0.0}

\title{Attention-Gated UNet Model for Semantic Segmentation of Brain Tumors and Feature Extraction for Survival Prognosis}

\author{
Rut Patel\\
Department of Information and Communication Technology\\
Pandit Deendayal Energy University\\
Gandhinagar, Gujarat, India\\
   \And
Snehal Rajput\\
Nirma University\\
Ahmedabad, Gujarat, India\\
     \And
Mehul S. Raval\\
Ahmedabad University\\
Ahmedabad, Gujarat, India\\
       \And
Rupal A. Kapdi\\
Nirma University\\
Ahmedabad, Gujarat, India\\
\And
Mohendra Roy\\
Department of Information and Communication Technology\\
Pandit Deendayal Energy University\\
Gandhinagar, Gujarat, India\\
\texttt{mohendra.roy@ieee.org}
}

\begin{document}
\maketitle

\begin{abstract}
Gliomas, among the most common primary brain tumors, vary widely in aggressiveness, prognosis, and histology, making treatment challenging due to complex and time-intensive surgical interventions. This study presents an Attention-Gated Recurrent Residual U-Net (R2U-Net) based Triplanar (2.5D) model for improved brain tumor segmentation. The proposed model enhances feature representation and segmentation accuracy by integrating residual, recurrent, and triplanar architectures while maintaining computational efficiency, potentially aiding in better treatment planning. The proposed method achieves a Dice Similarity Score (DSC) of 0.900 for Whole Tumor (WT) segmentation on the BraTS2021 validation set, demonstrating performance comparable to leading models. Additionally, the triplanar network extracts 64 features per planar model for survival days prediction, which are reduced to 28 using an Artificial Neural Network (ANN). This approach achieves an accuracy of 45.71\%, a Mean Squared Error (MSE) of 108,318.128, and a Spearman Rank Correlation Coefficient (SRC) of 0.338 on the test dataset.
\end{abstract}



\section{Introduction}
\label{sec:introduction}
Gliomas are the most common primary brain malignancies in the central nervous system (CNS) that arise from glial cells. Gliomas can be classified as high-grade gliomas (HGG) or low-grade gliomas (LGG), with HGG (called glioblastoma) being more aggressive, fatal, and having more diffusive boundaries than LGG. In the United States, glioblastoma accounts for 25\% of newly diagnosed primary brain tumor cases \cite{mesfin2023}. Despite significant technological advancements, the median survival rate remains approximately ten months \cite{rajput2020review}.
These tumors vary in aggressiveness, prognosis, and histological composition, including subregions such as edematous lesions, necrotic cores, and active and nonactive-enhancing cores \cite{bakas2018identifying}. This inherent heterogeneity is also evident in their radio imaging, with each subregion displaying distinct intensity patterns across multiparametric imaging scans, indicating unique biological properties \cite{menze2014multimodal}. 

Magnetic Resonance Imaging (MRI) is extensively utilized for tumor examination due to its non-harmful nature, high contrast, and high resolution. Manually segmenting tumors in MRI is time-demanding and susceptible to subjective errors. Automation in segmentation would significantly benefit oncologists by facilitating early diagnosis and treatment planning.

The Brain Tumor Segmentation (BRATS) challenge \cite{bakas2018identifying,menze2014multimodal,bakas2017advancing} offers a comprehensive multiparametric MRI dataset along with a platform for evaluating and comparing segmentation techniques. The task involves segmenting the tumor into three regions: Whole Tumor (WT), Enhancing Tumor (ET), and Tumor Core (TC). Additionally, the challenge includes predicting brain tumor patients' survival days (SD) by integrating features derived from these tumor regions with clinical data. The SD is categorized into long-term, mid-term, and short-term survival groups.

\subsection{Recent advances in brain tumor segmentation and SD prediction}\label{review}
Magadza et al. \cite{magadza2023efficient} proposed modifications to the baseline nnU-Net \cite{isensee2021nnu} that was the winner of the BRaTS 2020 Challenge, replacing standard convolutions with separate depthwise convolutions and adding bottleneck units to reduce the number of parameters and adding a shuffle attention mechanism to improve segmentation performance. They achieved a Dice Similarity Coefficient (DSC) of 0.912 (WT), 0.848 (TC), and 0.792 (ET) on the BraTS 2020 validation dataset. Rajput et al. \cite{rajput2024triplanar} used a 2D triplanar ensemble-based approach, implementing squeeze and excitation techniques \cite{hu2018squeeze} along with modified spatial and channel attention mechanisms \cite{roy2018concurrent} to achieve DSC of 0.873 (WT), 0.778 (TC), 0.713 (ET). Rajput et al. \cite{rajput2023multi} also proposed an ensemble of 2D triplanar and planar networks employing multiple attention mechanisms, achieving DSC of 0.875 (WT), 0.782 (TC), and 0.699 (ET). Raza et al. \cite{raza2023dresu} suggested a 3D deep residual U-Net that overcomes the vanishing gradient problems by introducing residual blocks in U-Net while preserving low-level features and easing the model training. The obtained DSC are 0.866 (WT), 0.836 (TC), and 0.800 (ET). They also cross-validated the proposed model on the BraTS 2021 benchmark dataset (50 patients), achieving DSC of 0.860 (WT), 0.840 (TC), and 0.822 (ET). Metlek et al. \cite{metlek2023resunet+} also proposed a residual connection-based U-Net to combat vanishing gradient problems. They also added connection nodes between the encoder and decoder layers to prevent losing features and improve the segmentation performance, achieving DSC of 0.928 (WT), 0.919 (TC), and 0.931 (ET). Henry et al. \cite {henry2021brain} focused on a 3D approach for segmentation using one encoder and 3 decoders. They modified the existing No New-Net architecture \cite{isensee2019no} and added two additional decoders, focusing each decoder on a specific tumor region and fusing the outputs to obtain the final segmentation map. 
The obtained DSC are 0.923 (WT), 0.866 (TC), and 0.826 (ET). 

For the BraTS 2021 Challenge, Peiris et al. \cite{peiris2021reciprocal} tested a combination of a U-Net-like segmentation module and a convolutional adversarial network. The approach used 3D convolutions and a module to synthesize adversarial examples. They achieved DSC of 0.908 (WT), 0.854 (TC), and 0.814 (ET) on the validation set. Hatamizadeh et al. \cite{hatamizadeh2021swin} proposed a 3D Transformer-based U-Net architecture following similar principles to those of Vision Transformers \cite{dosovitskiy2020image} and Swin Transformers \cite{liu2021swin}. They replace UNETR's encoder \cite{hatamizadeh2022unetr} with the Swin Transformer encoder and implement shifted window attention to achieve DSC of 0.926 (WT), 0.885 (TC), and 0.858 (ET).

For SD prediction in the BraTS 2020 Challenge, Rajput et al. \cite{rajput2022glioblastoma} extracted radiomic features from predicted segmented outputs and applied recursive feature elimination to reduce the number of features to 20. They trained a gradient-boosting regressor model on the features to obtain metrics of 0.62 (Accuracy) and 141065.30 (MSE) on the validation set. Ali et al. \cite{ali2021glioma} used radiomics and image-based features to predict SD. They further scaled down the extracted feature set using Random Forest-Recursive Feature Elimination (RF-RFE) and concatenated the age of patients to it. They trained a Random Forest (RF) Regressor on the selected feature set to achieve metrics of 0.483 (Accuracy) and 105079.40 (MSE). Agravat et al. \cite{agravat20203d} extract statistical and radiomic features and concatenate the age of the patients with the feature set. They use an RF Regressor to predict the SD. They obtain metrics of 0.517 (Accuracy) and 116083.477 (MSE). Patel et al. \cite{patel2021segmentation} extract features from the segmentation network multiple times and fuse the feature sets. They dimensionally reduce the feature set and concatenate clinical features with gross total resection (GTR) status before passing it into a Cox proportional Hazards model. They achieved metrics of 0.655 (Accuracy) and 152467 (MSE). Rajput et al. \cite{rajput2023interpretable} employed an efficient 3D segmentation network and used an RF model for SD prediction, utilizing location-based and radiomics-based features. They reported an accuracy of 0.552, an MSE of 79826.24, and a Spearman rank coefficient (SRC) of 0.711 on the BraTS2020 validation dataset. 

Recently, 3D Unet-based ensemble segmentation techniques \cite{ronneberger2015u} have reached human-level accuracy with reduced inter-intra variability. However, as the model grows larger and deeper, the computational parameters of the architecture increase significantly, making the training process more complex and computationally intensive. Consequently, recent studies focus on "efficient" or "lightweight" segmentation networks, including 2D and 2.5D networks \cite{sundaresan2020brain,rajput2023multi,baheti2022leveraging,ds2024enhancing}.

This study introduces a tri-planar (2.5D) segmentation network enhanced with an attention-gated recurrent, residual architecture (R2U-Net) designed to advance efficient and accurate medical imaging analysis. The R2U-Net offers several advantages: the residual network mitigates the gradient vanishing problem. It facilitates the training of deeper architectures, while the recurrent network captures temporal relationships and allows iterative refinement of features, resulting in richer feature representations. The 2.5D approach further reduces computational complexity, enhances interpretability, simplifies data acquisition, and minimizes memory requirements. Additionally, the encoder of the proposed network is employed to extract richer features to predict the survival days (SD) of glioma patients.

The remainder of the paper is organized as follows: Section \ref{review} discusses advanced techniques for brain tumor segmentation and survival day prediction. Section \ref{experiment} outlines the specifics of the dataset and introduces the proposed approach for segmentation and SD prediction. Section \ref{conclusion} elaborates on the findings and interpretations. Section \ref{conclusion} wraps up the methodology, highlights its constraints, and suggests future avenues for research.

\section{Methods and Materials}\label{method}

The schematic diagram of the proposed methodology is depicted in Figure \ref{figure4}. Three identical 2D networks were employed for segmentation, each trained on a different planar view (sagittal, coronal, and axial), as depicted in Figure \ref{figure1}. The final segmentation is derived by averaging the probability maps produced by these three models. Further, we have utilized the BraTS2020 training set to train the segmentation network, while the entire BraTS2021 training and BraTS2021 validation set were used to validate the proposed segmentation model. Note that the BraTS 2020 dataset is a part of the BraTS 2021 dataset. The BraTS2020 training set was used for SD prediction, as SD prediction was no longer part of the BraTS2021 dataset. The details of the BraTS dataset are provided below:
\begin{figure}[H]
\includegraphics[width=1.05\linewidth]{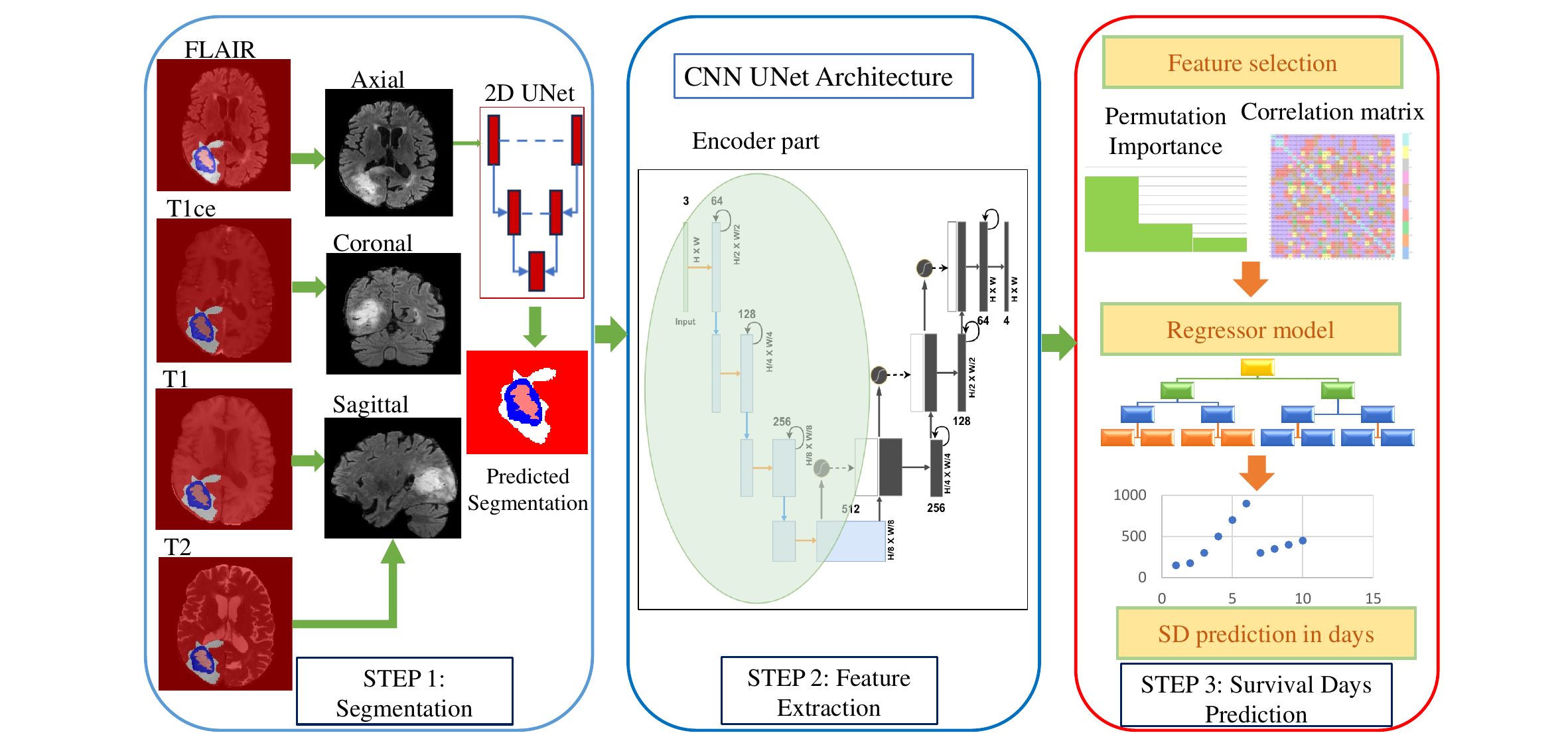}
 \caption{Schematic representation of proposed methodology for brain tumor segmentation and SD prediction. In step 1, a 2D UNet for each axes of the brain MRI is trained to segment the tumor (for each planner view). In step 2, the latent layer of the Encoder part of the 2D UNets (block shaded in green) is used to extract the features. In step 3, these features are analyzed and used to predict the survival days.}
    \label{figure4}
\end{figure}

\subsection{Datasets}
In this research, we employed the BraTS 2020 dataset (\cite{bakas2017advancing,menze2014multimodal,bakas2017segmentation}), which comprises 3D multiparametric MRI scans. The training data set comprises 369 samples from 19 institutions, each using different image acquisition protocols and MRI scanners. Expert neuroradiologists manually annotate the dataset to ensure accurate tumor segmentation. One to four specialist neurologists manually annotated actual labels. Details on the multiparametric MRI (mpMRI) data are provided in Table 1. 

For SD prediction, the BraTS2020 dataset includes information on 236 samples, which provides details on resection status, age, and survival days. The BraTS2021 validation set contains 219 mpMRI scans. The annotated labels of MRI scans are 1 for the necrotic and non-enhancing core, 2 for the edema region, 4 for the enhancing tumor, and 0 for the background voxels. The challenge organizers withheld the ground truth for the validation and test sets for later ranking of segmentation performance, with the test sets available only to the challenge participants. Each MRI modality has a volume size of 240$\times$240$\times$155 (width$\times$height$\times$depth per slice). Several pre-processing steps were performed by the BraTS organizers, including skull sampling and resampling to a voxel size of 1$\times$1$\times$1. 

\begin{table}
\centering
\caption{MRI modality and slice details about the BraTS2020 and BraTS2021 dataset \cite{menze2014multimodal}. A denotes axial, C denotes coronal, and S denotes sagittal.}\label{table0}
\centering\footnotesize
{\begin{adjustbox}{width=0.48\textwidth}
\begin{threeparttable}
\renewcommand{\arraystretch}{1.3}
\begin{tabular}{|p{1.07cm}|p{2.5cm}|p{1.8cm}|}
\hline
Modality & (3D / 2D) acquisition planes  &  Thickness of the voxel/slice\\\hline
T1W & 2D - A / S   & 1 – 6 \;mm\\\hline
T1ce & 3D & $1\;mm^3$\\
\hline
FLAIR & 2D - A / S / C & 2 – 6 \;mm\\\hline
T2  & 2D - A & 2 – 6\; mm\\\hline
\end{tabular}

\end{threeparttable}
\end{adjustbox}}
\end{table}

\subsection{Methodology for Segmentation}
\begin{figure}
    \centering \includegraphics[width=\linewidth]{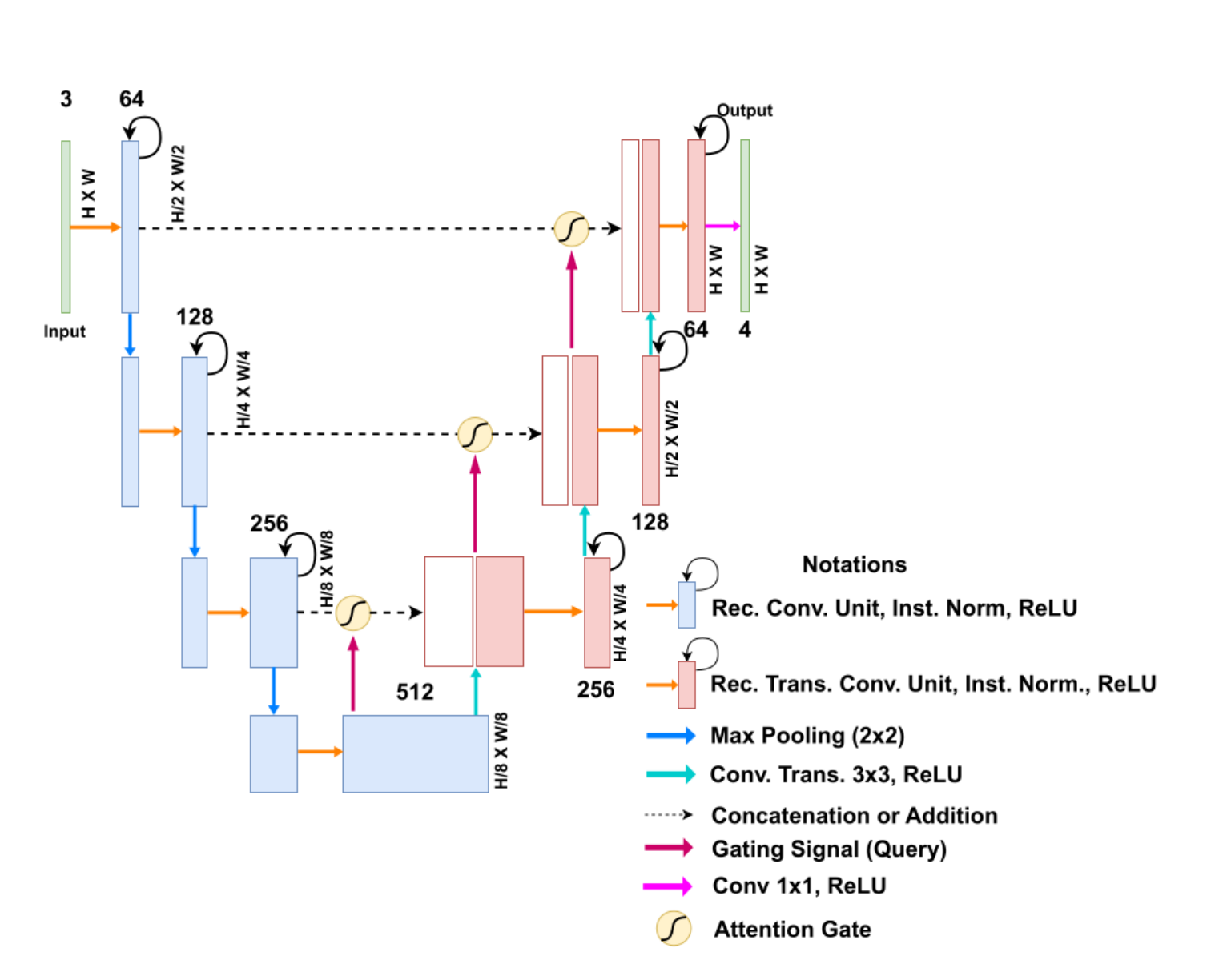}
    \caption{The structural diagram for 2D segmentation network.}
    \label{figure1}
\end{figure}

\begin{figure}
    \centering \includegraphics[width=\linewidth]{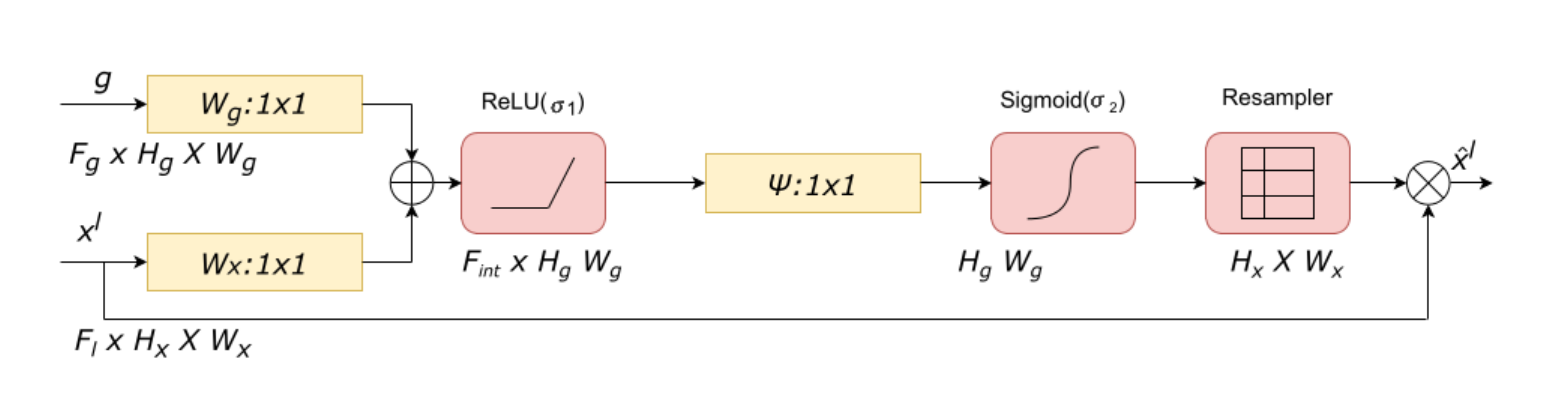}
    \caption{Schematic of the Attention-block used in the 2D UNet.}
    \label{figure2}
\end{figure}

The structural diagram of the proposed methodology for segmentation and prediction of SD is shown in Figures \ref{figure1} and \ref{figure3}. For segmentation, we implemented the 2D Attention-Gated R2U-Nets-based network \cite{alom2018recurrent,oktay2018attention}, shown in Figure \ref{figure2}. The network is a four-layer encoder-decoder architecture that incorporates recurring residual convolutional neural network units and attention gates within the traditional U-Net framework, as shown in Figure \ref{figure2}.  The initial number of filters at the encoder end is 64, gradually increasing to 512. The recurrent neural networks contain stacks of Recurrent Convolutional Layers (RCLs) whose states are computed over discrete time steps \cite{liang2015recurrent}. This recurrent mechanism enables the RCL to accumulate and refine features over time, which is particularly useful for tasks requiring context integration, such as object recognition. It also helps layers learn dependencies between features at different spatial locations, particularly useful for segmentation tasks requiring detailed boundary delineation. This dynamic interaction enhances the model's capacity to integrate contextual information, which is crucial for effective lesion segmentation in an image.  

In addition, a residual unit contains a residual mapping for learning residual functions $\mathcal{F}(x) + x$, and shortcut connections to address vanishing/exploding gradients and ease deep network training \cite{he2016deep}. The integration of RCL into RCNNs and residual units allows feature accumulation that improves feature representation \cite{alom2018recurrent}. At a given time step \(t\), the net input for the unit is calculated as:

\begin{equation}
Z_{ijk}(t) = \left(w^f \cdot u_{(i,j)}(t)\right) + \left(w^k \cdot x_{(i,j)}(t - 1)\right) + b_k
\end{equation}

where:

\begin{itemize}
    \item \(u_{(i,j)}(t)\): The current input at location \((i, j)\).
    \item \(x_{(i,j)}(t - 1)\): The output state from the previous time step, contributing to the recurrent computation.
    \item \(w^f\) and \(w^k\): Weights for the feedforward and recurrent connections, respectively.
    \item \(b_k\): A bias term for the feature map \(k\).
\end{itemize}

Down-sampling odd dimensions in an image can lead to complications when concatenating them to the decoder through skip connections, as up-sampling may result in even dimensions. For instance, an image dimension may be reduced from 19 to 9 during down-sampling. Still, during up-sampling, it would increase from 9 to 18, causing a dimension mismatch during concatenation in the skip connections. Therefore, to resolve this, the architecture is modified to incorporate slices of any dimension, using bilinear interpolation in attention blocks to handle the mismatch of dimensions while up-sampling the slice images. Images are up-sampled using transposed convolutions, enabling the capture of spatial relationships between pixels during the process. Due to the small batch size, Instance Normalization \cite{ulyanov2016instance} is employed instead of Batch Normalization (BN) \cite{ioffe2015batch}, as BN tends to degrade model performance with smaller batch sizes. BN degrades the model's performance in the case of smaller batch sizes \cite{wu2018group}. While up-sampling in the decoder, skip connections from the encoder help the model recover the spatial information of the down-sampled images. An attention gate is added at the skip connection at each level between the encoder and decoder's paths. At each layer, the attention gates receive the encoder's outputs, which contain rich spatial features, along with a gating signal from the deeper layers of the decoder. This allows the model to focus on salient features while suppressing irrelevant and noisy ones, using contextual information about regions of interest \cite{oktay2018attention}. This eliminates the need for separate object localization models for training and reduces the complexity of the model by passing only the features of importance to the subsequent decoder layers. At the last layer, the softmax function converts the outputs into probabilistic feature maps for the four tumor labels predicted by the model.

\subsection{Methodology for SD Prediction}

\begin{figure}[h]
\includegraphics[width=\linewidth]{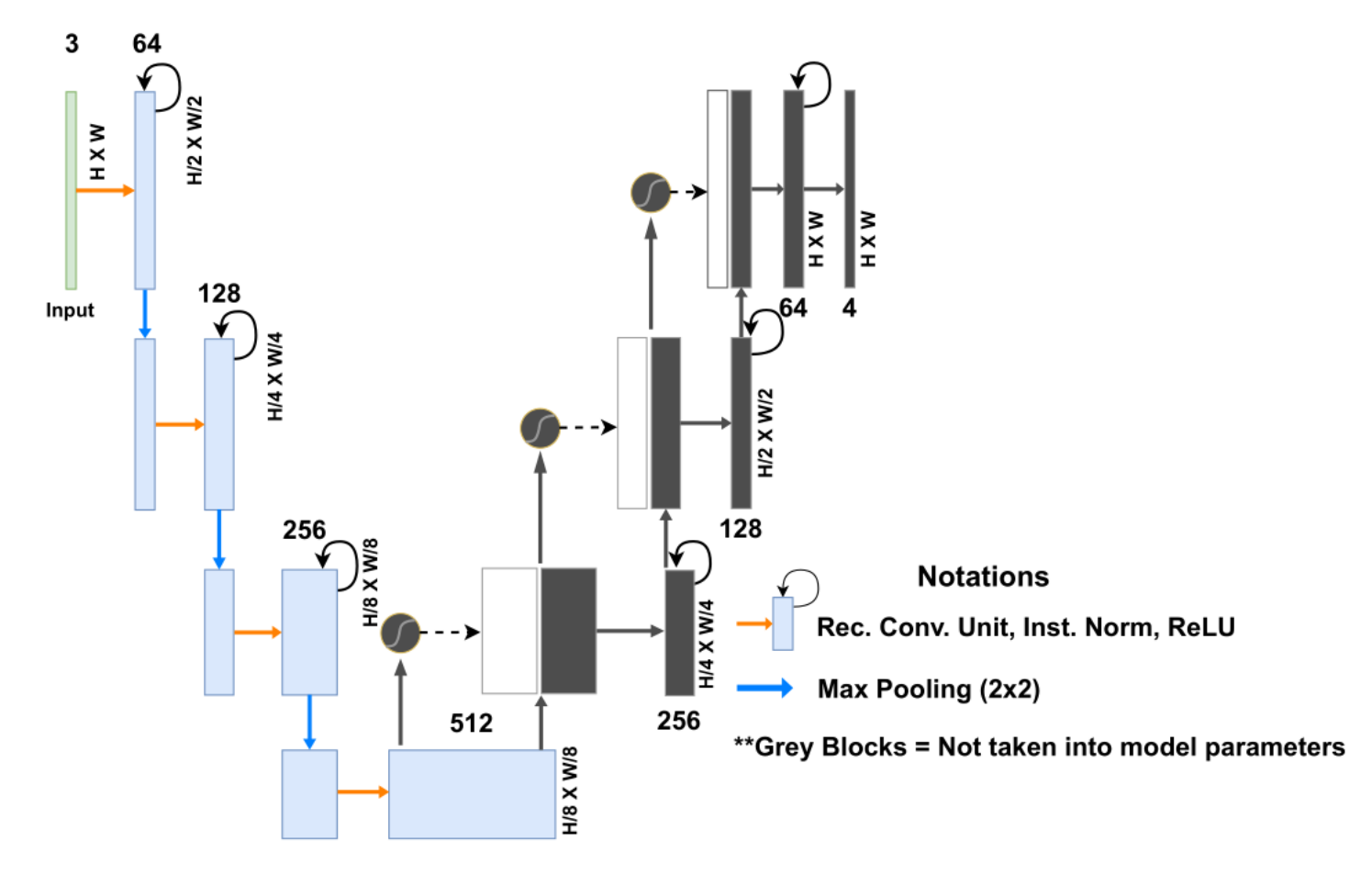}
    \caption{CNN network utilized for SD prediction. Here, only the encoder part is used to extract features, where the features are derived from the code layer of the encoder block (as shown in blue.) }
    \label{figure3}
\end{figure}

\begin{figure}[h]
\includegraphics[width=\linewidth]{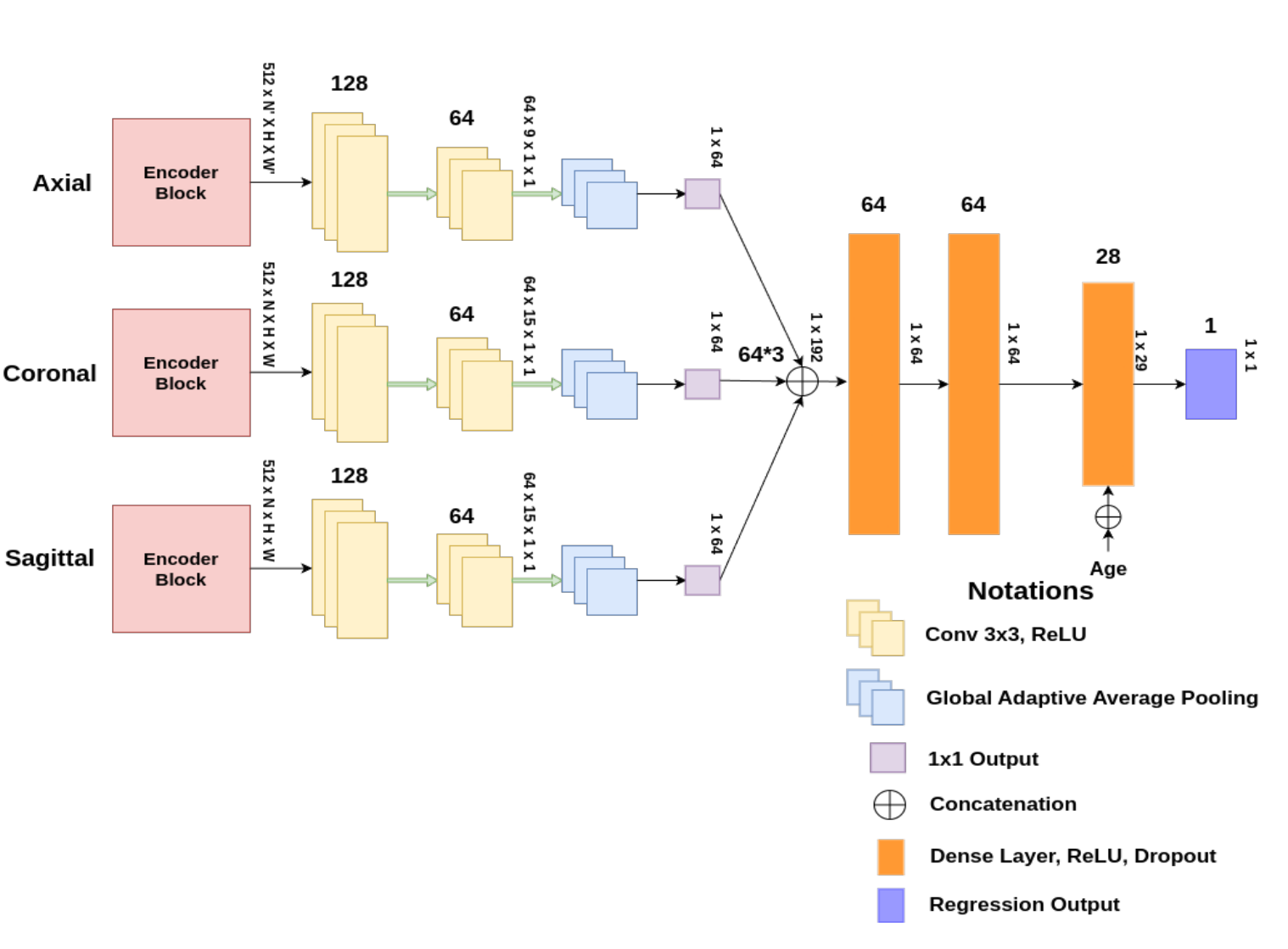}
    \caption{Schematic of Novel SD prediction methodology. The features from each of the encoder blocks of the 2D UNet of the axial, coronal, and sagittal planes are being fused to train an ANN (in orange color blocks) for SD prediction.}
    \label{figure3a}
\end{figure}

We have adopted a novel SD prediction method, which utilizes only the encoder part, as shown in Figure \ref{figure3}. The encoder consists of three-layered CNN blocks, as shown in Figure \ref{figure3}. We extract features from the bottleneck layer of the encoder, which captures most of the spatial information of slices passed to the R2U-Nets in a compressed feature vector space having 64 features for each axis using convolutional layers and global average pooling.  These features from each model are then concatenated along a common axis to represent a unified feature representation of 192 features. We empirically observed that the choice of 64 features, as layers with a higher number of neurons led to overfitting of the data, where the training loss function decreased significantly but performed worse on the validation dataset. Conversely, using fewer than 64 features was not optimal, since we observed the loss function not drop below a certain threshold, which is preferable for the model's learning capacity. The representation encodes multi-plane spatial details essential for SD prediction. The 192 features are fed into a 3-layer Artificial Neural Network (ANN).  The combined 192 features are fed into a 3-layer Artificial Neural Network (ANN) consisting of 2 layers having 64 neurons each and a single layer comprising 28 neurons to extract 28 key features and is combined with patients' age, totaling the features count to 29 which are then passed into the final layer containing a single neuron used to predict the SD. Dropout (rate: 0.3) and ReLU activation function are used in all the layers except the last, as shown in Figure \ref{figure3a}. 

\subsection{Loss Function}
For brain tumor segmentation, the Dice coefficient is calculated for each class, and its mean is used to determine the final Dice loss:

\begin{equation}
\text{Dice Loss} = 1 - \frac{1}{C} \sum_{c=1}^{C} 
\frac{2 \cdot \sum_{i} (p_{i,c} \cdot t_{i,c}) + \epsilon}
{\sum_{i} p_{i,c} + \sum_{i} t_{i,c} + \epsilon}
\end{equation}

Where:
\begin{itemize}
    \item \(p_i\): Predicted probability for the \(i\)-th voxel.
    \item \(t_i\): Target/ground truth value for the \(i\)-th voxel (binary or one-hot encoded).
    \item \(\epsilon\): A small constant to avoid division by zero.
    \item \(C\): Number of classes.
\end{itemize}

\subsection*{2. Focal Loss}
The Focal Loss is given by:

\[
\text{Focal Loss} = -\alpha \cdot (1 - p_t)^\gamma \cdot \log(p_t)
\]

For multiple classes:

\begin{equation}
\text{Focal Loss} = -\frac{1}{N} \sum_{i} \sum_{c=1}^{C} 
\alpha_c \cdot t_{i,c} \cdot (1 - p_{i,c})^\gamma \cdot \log(p_{i,c} + \epsilon)
\end{equation}

Where:
\begin{itemize}
    \item \(p_t = p_i\) if \(t_i = 1\), and \(p_t = 1 - p_i\) otherwise.
    \item \(\alpha\): Weighting factor for balancing the positive and negative samples.
    \item \(\gamma\): Focusing parameter to minimize the relative loss for well-classified examples (\(\gamma > 0\)).
    \item \(N\): Total number of voxels.
    \item \(p_{i,c}\): Predicted probability for class \(c\) at voxel \(i\).
    \item \(t_{i,c}\): Target/ground truth value (one-hot encoded) for class \(c\) at voxel \(i\).
    \item \(\epsilon\): A small constant to avoid \(\log(0)\).
\end{itemize}

\subsection*{3. Total Loss}
The total loss is the sum of the Dice Loss and Focal Loss from Equations 1 and 2:

\begin{equation}
    \text{Total Loss} = \text{Dice Loss} + \text{Focal Loss}
\end{equation}

\subsection{Performance Evaluation}
For segmentation, we utilized DSC, which measures the spatial overlap between two sets, such as predicted and ground truth segmentation in medical images. The ideal value for the DSC is 1, indicating perfect overlap. Hausdorff Distance (HD) quantifies the maximum distance between the boundary points of two sets, typically used to evaluate the spatial agreement between predicted and ground truth shapes in segmentation tasks. The ideal value for the HD is 0, indicating no boundary discrepancy.
Sensitivity is calculated as the ratio of true positives (TP) to the sum of true positives and false negatives (FN), which focuses on capturing true positives. whereas Specificity is computed as the ratio of true negatives (TN) to the sum of true negatives and false positives (FP), which ensures minimizing false positives. Mathematically, DSC and HD are defined as:  

\begin{equation}
    DSC = \frac{\textit{2TP}}{\textit{FP + 2TP + FN}}
\end{equation}

\begin{equation}
    HD = \max\, 
   \Bigl\{
    \underset{g\epsilon G}{S} \quad \underset{p\epsilon P}{I} \quad d\left(g,p\right), \quad \underset{p\epsilon P}{S} \quad \underset{g\epsilon G}{I} \quad d\left(g,p\right)
   \Bigr\}
 \end{equation}

where $FN$, $FP$, $TP$, $S$, and $I$ represent false negative, false-positive, true positive predictions supremum, and infimum, respectively, and $d(g, p)$ is the distance between points $g{\,\epsilon \,G}$ and points $p{\,\epsilon \,P}$.

For SD prediction, accuracy, Mean squared error (MSE), and 
Spearson correlation coefficient (SpearmanR) is used for performance evaluation.

\section{Experimentation}\label{experiment}
We have obtained the results using Python 3.7 \cite{van2009python} and PyTorch \cite{paszke2019pytorch} for training and validation purposes. The models were trained on a single Nvidia RTX A5000 GPU with 24 GB VRAM and 125GB RAM.

\subsection{Data Preprocessing and Augmentation}
We performed the following preprocessing on MRI scans: MRI scans often display uneven intensity distribution due to bias fields; therefore, we initially corrected for bias using the N4ITK package \cite{tustison2010n4itk}. Non-brain areas were also removed from all MRI modalities. Next, we excluded the top and bottom 1\% of intensities, regarded as outliers. Finally, intensity normalization was performed on each slice using mean and standard deviation values \cite{rajput2024triplanar,rajput2023multi}. All the images were feature-scaled to a range of 0 to 1. The dataset was meticulously reviewed, and unnecessary voxels were removed by cropping the slices along each axis: Sagittal (190×190), Coronal (190×140), and Axial (190×190). 

Further, we utilized the following data augmentation techniques for each slice before feeding it into the model to enhance robustness and mitigate overfitting. These techniques included horizontal flipping with a probability of 0.4, elastic transformation with a probability of 0.3, rotation with a probability of 0.4, shifting, scaling, and rotating the images with a probability of 0.3, as well as adding Gaussian noise with a probability of 0.2 and applying Gaussian blur with a probability of 0.2.

\subsection{Training}

During training, we used only the T1ce, T2, and FLAIR modalities of the sample images, combining them into a single multi-channel input before feeding them into the models. This approach offered several advantages, including leveraging complementary information for improved feature extraction, enhanced sensitivity and specificity in identifying tissue types and abnormalities, greater robustness against artifacts or low-quality data in individual modalities, and a more comprehensive context to support better decision-making \cite{lipkova2022artificial, huang2020fusion}.

We trained three distinct 2D models with the same architecture, one for each planar view (axial, coronal, and sagittal). For each model, 8 slices from the respective axis were stacked and used as input for each sample. The batch size of the mentioned inputs was set to 4 to accommodate more slices and capture more spatial context among the 2D slices. These slices could either be from the central part of the volume or randomly selected from the images, with the selection process being guided by a random index generated each time the model samples data. This ensures that the selection process is not always fixed and adds variability to the training process. 
  Adam optimizer \cite{kingma2014adam} with a learning rate of 0.00001 was used and the models were trained with a combination of Dice loss and Focal loss \cite{lin2017focal} throughout the training process. The sagittal model was trained for 1300 iterations, while the coronal and axial models were trained for 800 iterations each. We have observed that the sagittal model trains and its loss function converge, but exhibit a slower convergence rate than the coronal and axial models. Due to slow convergence, the sagittal model required more epochs to ensure optimal performance. This variation in the number of iterations between models was necessary to balance their training efficiency and convergence behavior.

For SD prediction, we split the BRaTS 2020 training dataset into a ratio of 85:15 for training and testing, respectively, using a fixed random seed and trained the CNN model on the training subset for 400 epochs using Adam optimizer \cite{kingma2014adam} with a learning rate of 0.0001. We used the mean squared loss (MSE) function to train the CNN model. 

\subsection{Inference}\label{discussions}

All 2D slices corresponding to each axis were input into their respective planar models for brain tumor segmentation. The output of each model consisted of a stack of slice-wise segmented maps, representing the segmentation results for each input slice. We then reconstructed the 3D segmentation maps for each sample by averaging the predictions from all the models and passing them through a softmax layer to predict the labels' probabilities. The segmented maps obtained were recalculated by converting the labels NET (1), ED (2), and ET (3) into WT (1$\cup$2$\cup$3), TC (1$\cup$3), and ET(3) to obtain the final 3D segmented maps for each sample. The performance of our proposed method on both the training and validation sets is presented in Table \ref{table2} and Table \ref{table3}. The method achieves high Dice Similarity Coefficient (DSC) scores, indicating strong segmentation accuracy across different tumor subregions. Specifically, the Whole Tumor (WT) achieves a DSC of 0.903, the Tumor Core (TC) attains 0.896, and the Enhancing Tumor (ET) scores 0.841. Additionally, the Hausdorff95 distance values—23.672 mm (WT), 11.852 mm (TC), and 11.099 mm (ET)—reflect precise boundary delineation, demonstrating robust segmentation capabilities. The specificity remains consistently high at 0.999 across all tumor regions, ensuring minimal false positives, while sensitivity values of 0.872 (WT), 0.874 (TC), and 0.806 (ET) indicate reliable detection of tumor structures. Furthermore, as shown in Table IV, our method achieves state-of-the-art results, outperforming leading approaches in DSC and Hausdorff95 distance, particularly for whole tumor (WT) and tumor core (TC) segmentation.

For SD prediction, we tested the data on the test subset. We extracted the features from the bottleneck layer of all three R2U-Net models and passed them into each R2U-Net's outputs through a 2-layered CNN to extract further features. The extracted features were then passed through a layer of global-average pooling to bring the features to a uniform dimension of 1$\times$1 among all the axes. All the features from the three models were concatenated together before being passed into a 3-layer ANN to predict the survival days.

\begin{table}
\caption{Performance on BraTS2021 training set.}\label{table2}
\centering\large
{\begin{adjustbox}{width=0.48\textwidth}
\begin{threeparttable}
\renewcommand{\arraystretch}{1.5}
\begin{tabular}{|p{1.4cm}|p{1.8cm}|p{2.6cm}|p{2cm}|p{2cm}|}
\hline
Tumor lesion & DSC & Hausdorff95 distance (mm) & Specificity & Sensitivity \\\hline
WT & 0.903 & 23.672 & 0.999 & 0.872 \\\hline
TC & 0.896 & 11.852 & 0.999 & 0.874 \\\hline
ET & 0.841 & 11.099 & 0.999 & 0.806 \\\hline
\end{tabular}
\end{threeparttable}
\end{adjustbox}}
\end{table}

\begin{table}
\caption{Performance on BraTS2021 validation set.}\label{table3}
\centering\large
{\begin{adjustbox}{width=0.48\textwidth}
\begin{threeparttable}
\renewcommand{\arraystretch}{1.5}
\begin{tabular}{|p{1.4cm}|p{2cm}|p{2.5cm}|p{2cm}|p{2cm}|}
\hline
Tumor lesion & DSC & Hausdorff95 distance (mm) & Specificity & Sensitivity \\\hline
WT & 0.900 & 05.232 & 0.999 & 0.889 \\\hline
TC & 0.824 & 12.001 &  0.999 & 0.795 \\\hline
ET & 0.775 & 18.422 &  0.999 & 0.756 \\\hline
\end{tabular}
\end{threeparttable}
\end{adjustbox}}
\end{table}

\begin{table}
\caption{Comparison of performance metrics for tumor regions across leading models on the BraTS2021 validation set. Best scores are in boldface.}
\centering
\resizebox{0.5\textwidth}{!}{%
\begin{threeparttable}
\renewcommand{\arraystretch}{1.5}
\begin{tabular}{|c|c|c|c|c|c|c|}
\hline
Method & \multicolumn{3}{c|}{DSC} & \multicolumn{3}{c|}{Hausdorff95 distance (mm)} \\
\hline
 & WT & TC & ET & WT & TC & ET \\
\hline
Maurya et al. \cite{maurya2021brain} &  0.888 &  0.781 & 0.761 & 7.51 & \textbf{13.360} & 23.88 \\ \hline
Yogananda et al. \cite{bangalore2021disparity} & \textbf{0.900} &
0.800 & 0.790 & \textbf{5.220} & 19.180 & 19.990 \\
\hline
Shi et al. \cite{shi2021ensemble} & 0.891 & 0.738 & \textbf{0.819} & 5.772 & 16.874 & \textbf{16.628} \\ \hline
Marndi et al. \cite{marndi2021brain} & 0.880 & 0.749 & 0.783 & NA & NA & NA \\
\hline
\rowcolor{atomictangerine!80!}
Proposed Method & \textbf{0.900} & \textbf{0.824} & 0.775 & 5.232 & 12.001 & 18.422 \\
\hline
\end{tabular}
\end{threeparttable}
}
\end{table}

\begin{figure}
    \centering

    \subfloat[]{%
        \includegraphics[width=0.3\linewidth]{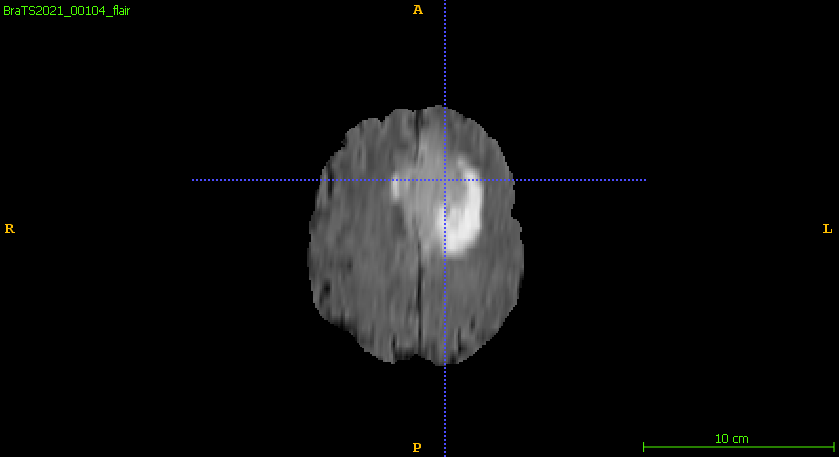}%
        \label{fig:sub01}
    }
    \hfill
    \subfloat[]{%
        \includegraphics[width=0.3\linewidth]{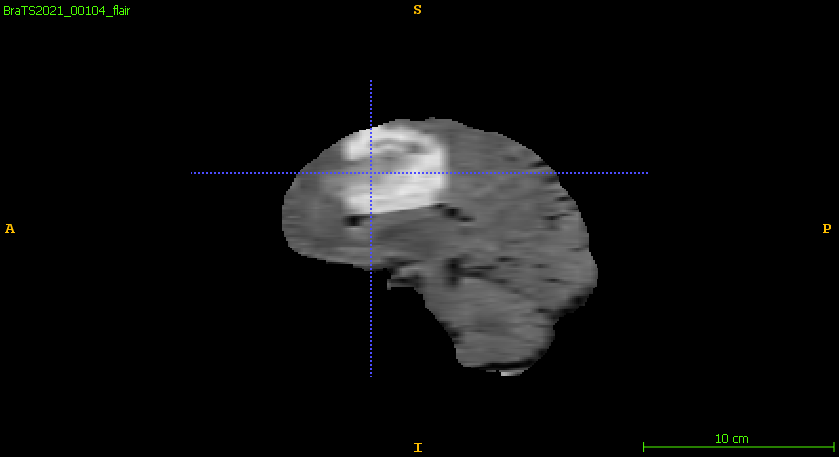}%
        \label{fig:sub02}
    }
    \hfill
    \subfloat[]{%
        \includegraphics[width=0.3\linewidth]{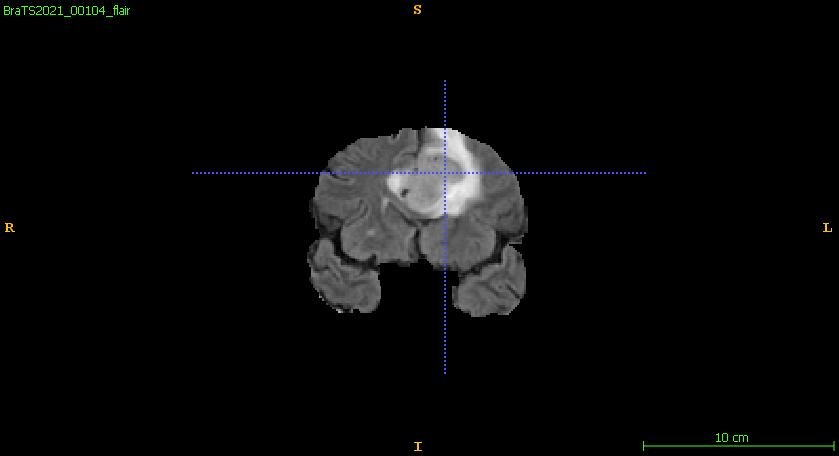}%
        \label{fig:sub03}
    }

    \vspace{0.3cm}

    \subfloat[]{%
        \includegraphics[width=0.3\linewidth]{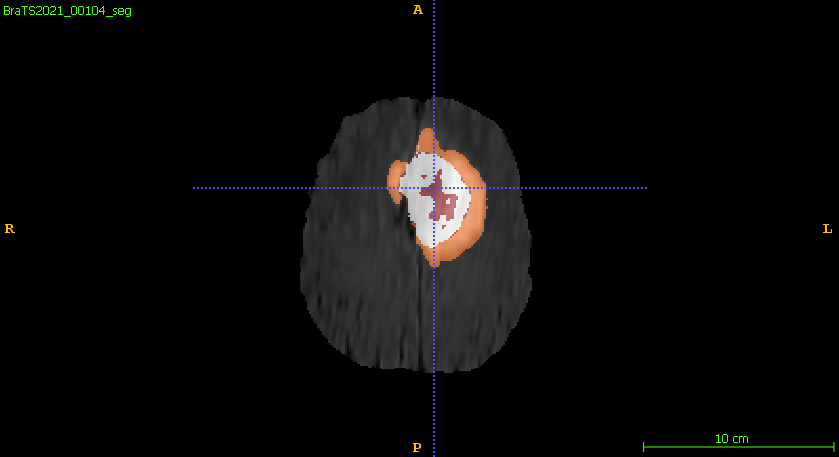}%
        \label{fig:sub1}
    }
    \hfill
    \subfloat[]{%
        \includegraphics[width=0.3\linewidth]{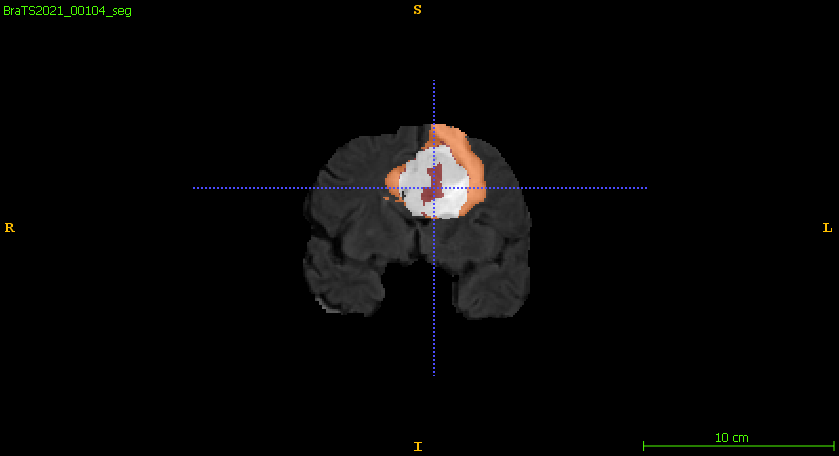}%
        \label{fig:sub2}
    }
    \hfill
    \subfloat[]{%
        \includegraphics[width=0.3\linewidth]{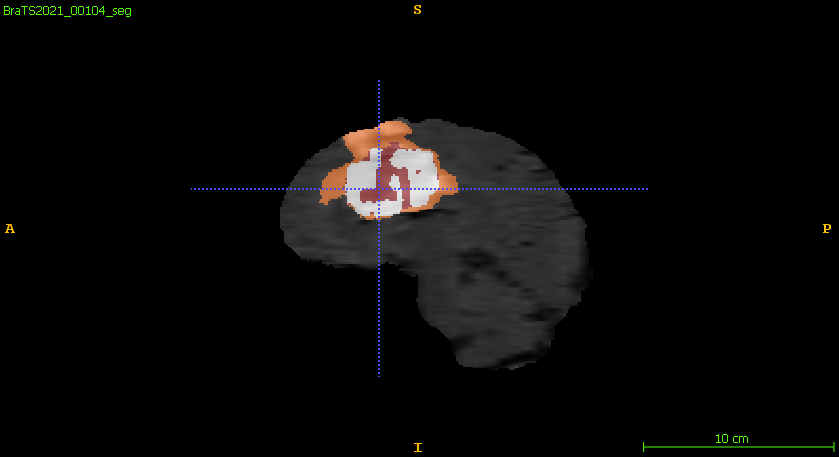}%
        \label{fig:sub3}
    }

    \vspace{0.3cm}

    \subfloat[]{%
        \includegraphics[width=0.3\linewidth]{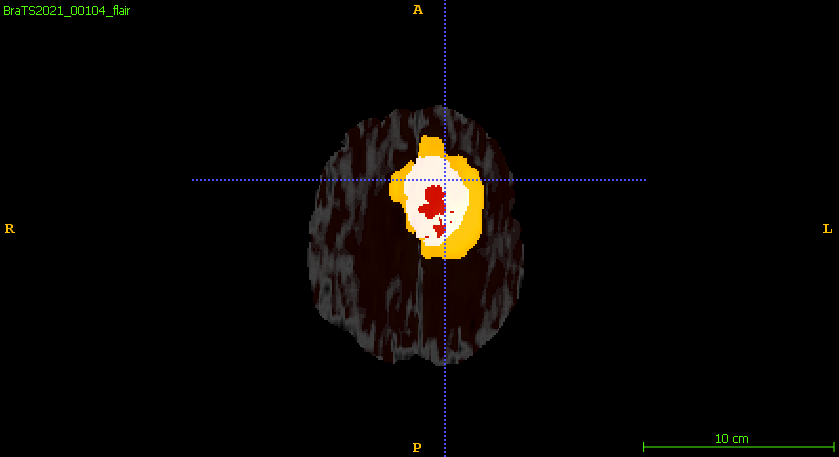}%
        \label{fig:sub4}
    }
    \hfill
    \subfloat[]{%
        \includegraphics[width=0.3\linewidth]{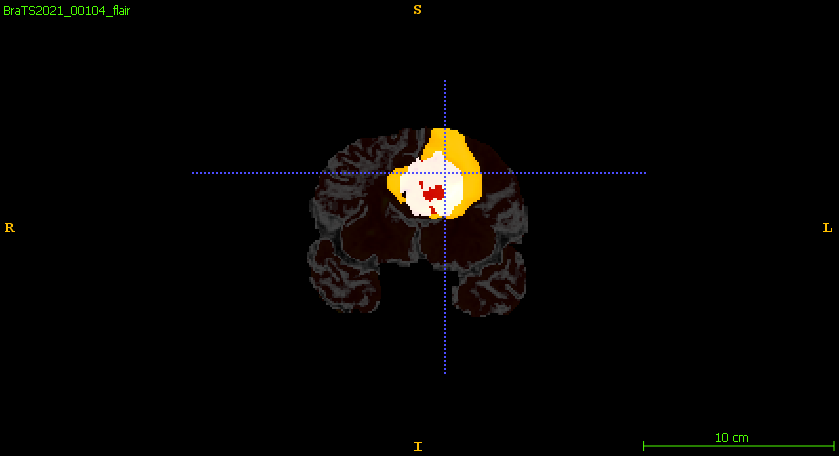}%
        \label{fig:sub5}
    }
    \hfill
    \subfloat[]{%
        \includegraphics[width=0.3\linewidth]{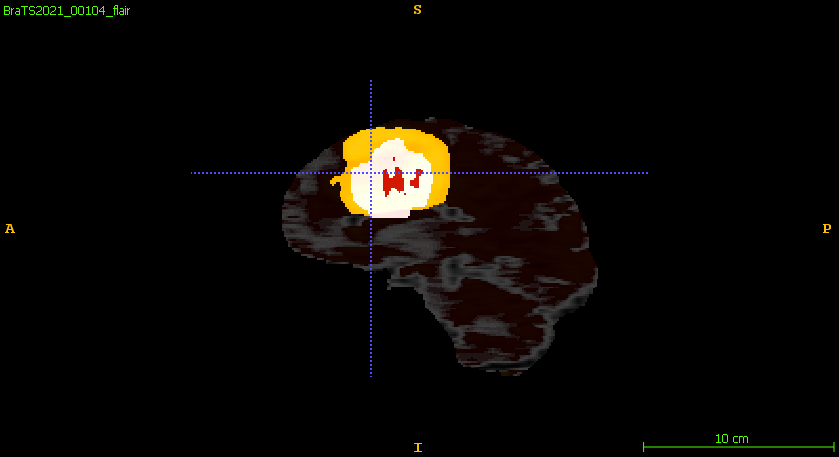}%
        \label{fig:sub6}}
\caption{Qualitative comparisons include (a–c) input FLAIR images with ground truth overlaid across axial, sagittal, and coronal views, (d–f) ground truth across the planar views, and (g–i) the corresponding predicted segmentation labels from the training set for each view.}\label{figure5}
\end{figure}

Furthermore, Figure \ref{figure5} compares the ground truth on different planar views and the predicted segmentation labels corresponding to those views in the training set. The segmentation network demonstrates robust performance, accurately capturing the intricate details of tumor boundaries and structures across all planar views. The achieved DSC for the samples shown is 0.943 for ET, 0.957 for WT, and 0.949 for TC, reflecting the network's high precision and reliability in segmenting different tumor regions. These results highlight the model's effectiveness in generating segmentation outputs closely aligned with the ground truth.

Table \ref{table5} shows the performance metrics for SD prediction from the BraTS2020 dataset. Similarly, Table \ref{table6} compares the performance with the leading BraTS models on the BraTS2020 dataset. Our SD predictor model has performed mediocre compared to other leading models with an accuracy of 45.71\%, MSE of 108318.128, and SpearmanR of 0.338. Directly utilizing deep features for SD prediction overfits, suggesting potential instability in the trained models. A similar observation has been reported in the study \cite{starke2020integrative}.

\begin{table}
\caption{Performance metrics for SD prediction from the BraTS2020 dataset. }\label{table5}
\centering\large
{\begin{adjustbox}{width=0.48\textwidth}
\begin{threeparttable}
\renewcommand{\arraystretch}{1.5}
\begin{tabular}{|p{3.5cm}|p{2.5cm}|p{2.5cm}|p{2.5cm}|}
\hline
Dataset & MSE & SpearmanR & Accuracy \\\hline
Training (85\%) & 51742.430 & 0.871 & 69.50\% \\\hline
Validation (15\%) & 108318.128 & 0.338 & 45.71\% \\\hline
\end{tabular}
\end{threeparttable}
\end{adjustbox}}
\end{table}

\begin{table}[H]
\caption{Performance comparison with leading models for SD prediction from the BraTS2020 dataset.}\label{table6}
\centering\large
{\begin{adjustbox}{width=0.48\textwidth}
\begin{threeparttable}
\renewcommand{\arraystretch}{1.5}
\begin{tabular}{|p{4cm}|p{2.5cm}|p{2cm}|p{2cm}|}
\hline
Dataset & MSE & SpearmanR & Accuracy \\\hline
\rowcolor{atomictangerine!80!}
Proposed method & 108318.128 & 0.338 & 45.71\% \\\hline
Rajput et al. \cite{rajput2023interpretable} & 079826.24 & 0.711 & 55.20\%\\\hline
Mckinley et al. \cite{mckinley2020uncertainty}  & 098704.66 & 0.253 & 41.40\%\\\hline

Bommineni et al. \cite{Bommineni2021} & 093859.54 & 0.280 & 37.90\% \\\hline
Asenjo et al. \cite{MartiAsenjo2021} & 122515.80 & 0.130 &52.00\% \\\hline
\end{tabular}
\end{threeparttable}
\end{adjustbox}}
\end{table}

\section{Conclusion, limitation and Future Work}\label{conclusion}
In this work, we introduced a modified Attention Gated R2U-Net model, derived from the R2U-Net architecture, specifically designed for brain tumor segmentation using MRI scans across multiple planes (sagittal, coronal, and axial). The model processes 2D slices and integrates the information from these slices to reconstruct 3D tumor regions, effectively addressing the limitations of 2D segmentation in capturing the geometric complexities of tumors. The proposed network demonstrated robust performance compared to 2D, 2.5D, and 3D models.

A multi-planar framework guarantees the integration of features drawn from different encoder blocks corresponding to distinct orientations (sagittal, coronal, and axial). This endows the model with the capability of capturing both the detailed spatial and contextual information necessary for accurate depiction of tumor subregions, such as the whole tumor, the tumor core, and the enhancing tumor. The results on the BraTS2021 dataset gave competitive DSC as that of some advanced methods and thus validate our model regarding segmentation efficacy and computational efficiency, having obtained DSC of 0.900, 0.824, and 0.775 for WT, TC, and ET, respectively.

Further, we present a novel method for predicting the SD of the patients. Our proposed method includes directly taking the outputs from R2U-Net's encoder's bottleneck layer and passing them through an architecture comprising of advanced pooling and convolutional techniques with a Fully Connected Neural Network (FCNN) at the end to predict the OS of the patient, bypassing the entire decoder's computation. We acquired an MSE of 108318.128 and a classification accuracy of 45.71\% on the validation split of the BRaTS 2020 training dataset.  This would complement the interpretability of the model, consequently reducing computation redundancy.

While the DSC achieved competitive results on the BraTS2021 dataset, our approach has certain limitations. Specifically, we trained our proposed segmentation network on the BraTS2020 dataset because our target was also SD prediction. We validated the network on the BraTS2021 dataset due to the discontinuation of BraTS2020 validation on the online evaluation platform. In addition, the BraTS ground truth of the validation set is not publicly available. Second, the segmentation model is still complex and requires considerable computation, which can be optimized further for clinical utility. Lastly, since it is a triplanar network, it lacks access to complete depth information; therefore, its ability to capture intricate volumetric features may be limited compared to fully 3D networks.

However, exploration of more efficient and lightweight triplanar or 3D models for brain tumor segmentation remains an area for future research. Combining deep learning features with radiomic features for SD prediction may enhance predictive accuracy. Furthermore, integrating multiomics data, including genetic, transcriptomic, proteomic, metabolomic, and epigenomic information, could further refine SD prediction performance.

\section*{Acknowledgment}
The authors thank M. Roy for facilitating the seed grant No. $ORSP/R\&D/PDPU/2019/MR-/RO051$ from PDEU (for the computing facility), the core research grant No. $CRG/2020/000869$ from the Science and Engineering Research Board (SERB), Government of India, and the project grant No. $GUJCOST/STI/2021-22/3873$ from GUJCOST, Government of Gujarat, India.

\bibliographystyle{unsrt}
\bibliography{references}

\end{document}